\documentclass[letterpaper, 10 pt, conference]{ieeeconf}

\IEEEoverridecommandlockouts
\usepackage{graphicx}
\usepackage{amsmath}
\usepackage{amssymb}
\usepackage{booktabs}
\usepackage{multirow}
\usepackage{array}
\usepackage{cite}
\usepackage{url}
\usepackage{placeins}

\title{\LARGE \bf
From a Single RGB Image to Robot Execution: Metric Scene Alignment and Contact-Guided Retargeting
}

\author{
Zhenxuan Ding$^{1}$, Yu Huang$^{2}$, Xinmiao Wang$^{3}$,\\
Chenghao Sun$^{3}$, Ziyang Dong$^{3}$, and Wenhao Cui$^{3,*}$%
\thanks{$^{1}$Beijing University of Chemical Technology; $^{2}$National University of Defense Technology; $^{3}$Noetix Robotics. $^{*}$Corresponding author: Wenhao Cui.}%
}

\begin{document}

\maketitle
\thispagestyle{empty}
\pagestyle{empty}

\begin{abstract}
Extending humanoid interaction generation to new environments remains limited by the need for prebuilt 3D scene assets. Reconstructing scenes directly from single RGB images offers a way to overcome this limitation, but introduces a key challenge in maintaining geometric consistency throughout motion generation and transfer. Visually plausible human motions may be misaligned with the reconstructed scene, while retargeting can introduce additional support errors due to differences between human and robot embodiments. 
We present a unified pipeline that transforms a single RGB image into a humanoid motion reference. It reconstructs scene geometry from the input image, generates a human interaction conditioned on the reconstructed scene, and retargets the motion to a humanoid robot. To maintain geometric consistency throughout this process, we incorporate two complementary refinement stages before and after embodiment transfer. Before retargeting, Joint Metric Interaction Refinement (JMIR)  refines human motion alignment with the registered metric scene while preserving the generated interaction structure. Following retargeting, Contact Guided Residual Retargeting (CGRR) corrects robot specific support errors to restore the intended support relationships. Together, these stages produce motion references that are consistent with the reconstructed scene and can be followed by multiple pretrained humanoid trackers without scene specific policy fine tuning. Compared with GRAIL, our method reduces refinement time from 480~s to 87~s, yielding a speedup of 5.5$\times$. The complete pipeline achieves a success rate of 89\% in simulation and 70\% across 50 physical trials on the Unitree G1, demonstrating its effectiveness in generating executable humanoid motion references from single RGB images.

\end{abstract}

\section{INTRODUCTION}
Recent advances in humanoid motion tracking have enabled the execution of diverse reference motions~\cite{gmt,sonic}, while motion retargeting has improved the transfer of human motion across robot embodiments~\cite{gmrpaper}. Despite this progress, generating suitable references for interaction with a specific environment remains a separate challenge. In interactions that depend on environmental support, motion quality is determined not only by the body trajectory but also by its geometric relationship with the scene. A reference must therefore preserve the intended support contacts while respecting the geometry of both the environment and the robot. Motion tracking alone does not resolve inconsistencies in these scenarios. Generative human object interaction (HOI) models offer a promising basis for constructing such references. HOI-Diff~\cite{hoidiff} and CHOIS~\cite{chois} generate coordinated human and object motions from language and geometric conditions, while DAViD~\cite{david} models dynamic object affordances using pretrained video diffusion models. These methods provide valuable human interaction priors, but do not directly address the geometric and kinematic requirements of a target humanoid robot. GRAIL~\cite{grail} advances toward robot execution by integrating interaction generation with humanoid retargeting and policy learning. However, its generation process assumes a predefined 3D scene with known object geometry and camera parameters, limiting its application to a new environment. 
\begin{figure*}[t]
\centering
\includegraphics[width=0.97\textwidth]{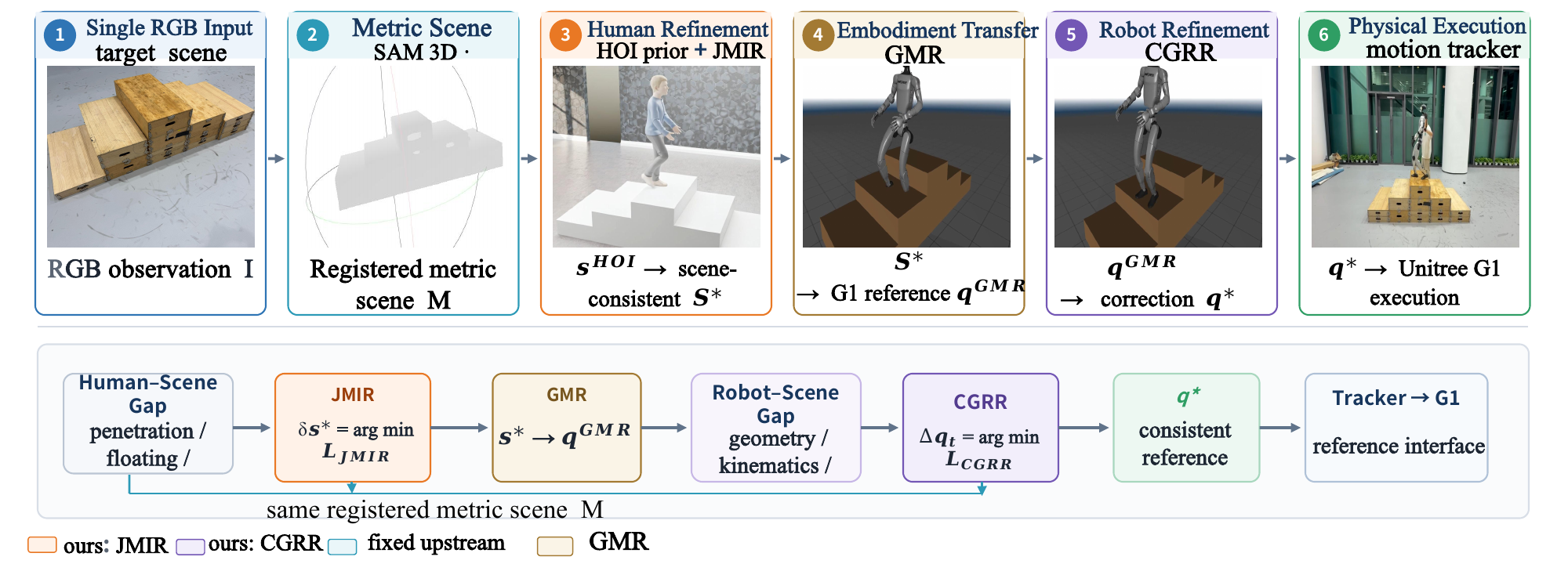}
\caption{Overview of the single-image-to-humanoid workflow. From a single RGB image, the pipeline reconstructs a registered metric scene and generates a scene-conditioned human interaction. JMIR first corrects human--scene inconsistency, GMR transfers the refined motion to Unitree G1, and CGRR then corrects the residual robot--scene support mismatch to produce an executable robot reference.}
\label{fig:pipeline}
\end{figure*}

Direct reconstruction from a single RGB image provides an alternative to preparing the target scene assets in advance, but it introduces challenges in maintaining geometric consistency during motion generation and transfer. The generated human interaction may be semantically appropriate yet misaligned with the reconstructed scene, resulting in penetration or loss of support contact. Even after this mismatch is corrected, retargeting can introduce additional support errors due to differences in body proportions, contact geometry, and joint limits. These two sources of error motivate complementary refinement before and after embodiment transfer: the first stage aligns the human interaction with the scene and preserves its structure, while the second uses the target robot geometry and kinematics to restore the intended support contacts.

In this paper, we present a unified pipeline that generates humanoid motion references from a single RGB image for interactions that rely on environmental support. As shown in Fig.~\ref{fig:pipeline}, we first reconstruct scene geometry using SAM 3D Objects~\cite{sam3d} and generate a human interaction conditioned on the reconstructed scene. To maintain geometric consistency before embodiment transfer, we introduce Joint Metric Interaction Refinement (JMIR), which aligns the generated human interaction with the registered metric scene while preserving its interaction structure. The refined motion is then transferred to the target humanoid through General Motion Retargeting (GMR)~\cite{gmrpaper}. Following retargeting, Contact Guided Residual Retargeting (CGRR) restores the intended support contacts through a bounded residual correction to the lower body. By separating scene alignment from robot support correction, we address geometric errors before and after embodiment transfer while retaining existing motion generation and retargeting components.
The resulting references can be used by pretrained humanoid motion trackers without scene specific policy fine tuning. We evaluate their compatibility with MimicLite~\cite{mimiclite}, GMT~\cite{gmt}, and SONIC~\cite{sonic}, and validate the complete workflow in simulation and on a physical Unitree G1 robot. Under matched evaluation settings, our method reduces refinement time from 480~s to 87~s compared with GRAIL. The complete pipeline achieves a success rate of 89\% in simulation and 70\% across 50 real robot trials.

Our main contributions are summarized as follows.
\begin{itemize}
\item \textbf{A one-RGB-image-to-humanoid pipeline.} We integrate scene reconstruction, human interaction generation, and motion retargeting into a unified pipeline for interactions that rely on environmental support, without requiring prebuilt assets of the target scene.
\item \textbf{Geometric refinement across embodiment transfer.} We introduce JMIR to refine human–scene geometric alignment while preserving the generated interaction structure, and CGRR to correct robot specific support errors after retargeting. Together, these stages address distinct geometric errors before and after embodiment transfer.
\item \textbf{Experimental validation in simulation and on real robots.} Experiments in simulation and on the Unitree G1 demonstrate that our pipeline generates executable humanoid motion references with reduced refinement time. We further demonstrate compatibility with multiple pretrained motion trackers without scene specific policy fine tuning.

\end{itemize}

\section{RELATED WORK}
\subsection{Human Interaction Generation and Visual Reconstruction}
Human motion models provide priors for interaction generation~\cite{mdm}. HOI-Diff~\cite{hoidiff} and CHOIS~\cite{chois} synthesize coordinated human and object motions from language and geometric conditions, while DAViD~\cite{david} uses pretrained video diffusion models to capture dynamic object affordances. These methods model human interactions, but their geometric validity does not directly carry over to robots with different body structures.
GRAIL~\cite{grail} connects interaction generation with humanoid retargeting and policy learning using predefined 3D assets and video priors. VideoMimic~\cite{videomimic} instead reconstructs human motion and the environment jointly from monocular demonstration videos. Both approaches require scene assets or human demonstrations, adding preparation effort when adapting to new environments.
Our method addresses this limitation by reconstructing the scene from a single RGB image using SAM 3D Objects~\cite{sam3d} and refining the generated interaction to improve geometric consistency.

\subsection{Motion Retargeting and Contact Preservation}
Motion retargeting transfers human motion across embodiments, but preserving body motion alone does not ensure valid environmental contacts. GMR~\cite{gmrpaper} provides a general retargeting framework, while OmniRetarget~\cite{omniretarget} preserves relationships among the body, terrain, and objects through an interaction mesh under kinematic constraints. KinoDynamic Motion Retargeting (KDMR)~\cite{kinodynamic} incorporates rigid body dynamics and contact complementarity constraints into trajectory optimization, using ground reaction force measurements to recover contact patterns. Our method retains GMR, with JMIR aligning the human interaction to the scene before retargeting and CGRR correcting residual robot support errors afterward, without full dynamic trajectory optimization.

\subsection{Humanoid Motion Tracking and Interaction Control}
Physics based motion imitation methods such as PHC~\cite{phc} enable simulated humanoids to track diverse motions and recover from failures. GMT~\cite{gmt} and SONIC~\cite{sonic} support general motion tracking on physical robots, while BeyondMimic~\cite{beyondmimic} combines tracking with guided diffusion for task conditioned control.
For interaction tasks, InterMimic~\cite{intermimic} uses teacher policies to refine imperfect motion capture demonstrations before distilling interaction skills into a shared policy. ResMimic~\cite{resmimic} learns residual policies on top of a pretrained tracker to improve object interaction, while SceneBot~\cite{scenebot} conditions tracking on reference motions and contact labels. These approaches address interaction requirements through policy learning or contact conditioning. Our work focuses on geometric refinement of references before tracking and evaluates their execution with multiple pretrained humanoid trackers without scene specific policy fine tuning.

\section{METHOD}
\subsection{Problem Formulation and Pipeline Overview}
Let $t\in\{1,\ldots,T\}$ index the motion frames, where $T$ is the sequence length. We denote the input RGB image by $I$, the registered metric scene by $M$, the generated human interaction by $S^{\mathrm{HOI}}_{1:T}$, the JMIR-refined interaction by $S^{\star}_{1:T}$, the initial Unitree G1 reference obtained after GMR by $q^{\mathrm{GMR}}_{1:T}$, and the final scene-consistent robot reference by $q^{\star}_{1:T}$.
Given $I$, a fixed SAM 3D module reconstructs the visible scene geometry and provides an estimated metric-scale 3D representation, $\widetilde{M}=F_{\mathrm{SAM3D}}(I)$~\cite{sam3d}. The reconstructed geometry is rigidly registered to the common interaction/robot coordinate frame to obtain $M$. SAM 3D is treated as a fixed upstream reconstruction module and its parameters are not optimized in our framework.

A fixed upstream HOI generator produces $S^{\mathrm{HOI}}_{1:T}$ conditioned on the reconstructed scene. The generated interaction contains the human root motion and articulated body configuration over time. From its interaction representation, we extract $\mathcal{C}_h$, the active human--scene interaction-point index set over frames, and $\mathcal{E}_t$, the interaction-edge set at frame $t$. These sets are fixed before JMIR. Specifically, $\mathcal{C}_h$ identifies the human points used for metric consistency, whereas $\mathcal{E}_t$ preserves their relative human--scene structure during refinement. The same registered scene $M$ is used by both refinement stages. Fig.~\ref{fig:pipeline} summarizes the complete workflow.

\subsection{Two-Stage Consistency Refinement Across Embodiment Transfer}
The proposed refinement is organized around two distinct sources of geometric inconsistency separated by embodiment transfer. Before retargeting, the generated human interaction may violate the reconstructed scene geometry through penetration, floating contact, or inaccurate relative placement. We refer to this as the \emph{human--scene gap} and correct it in the human representation using JMIR. After the refined interaction is transferred to Unitree G1 through GMR, differences in body proportions, sole geometry, joint limits, and kinematics can introduce a new support mismatch even when the human motion is already scene-consistent. We refer to this as the \emph{robot--scene gap} and correct it in the robot representation using CGRR. GMR therefore forms the embodiment-transfer boundary between the two refinement stages. The staged design preserves the generated interaction prior and reduces optimization cost compared with global joint optimization.

\subsubsection{Human--Scene Consistency Refinement}
JMIR corrects the human--scene gap by optimizing a residual sequence $\delta S_{1:T}$ around the generated interaction,
\begin{equation}
S_t^{\star}=S_t^{\mathrm{HOI}}+\delta S_t
\end{equation}
Rather than introducing the individual penalties first, we formulate JMIR directly as:
\begin{equation}
\delta S_{1:T}^{\star}
=
\arg\min_{\delta S_{1:T}}
\mathcal{L}_{\mathrm{JMIR}}
\end{equation}
with the total objective $\mathcal{L}_{\mathrm{JMIR}}$, which is defined as:
\begin{equation}
\mathcal{L}_{\mathrm{JMIR}}
=
\lambda_m\mathcal{L}_{\mathrm{metric}}
+\lambda_i\mathcal{L}_{\mathrm{inter}}
+\lambda_r\mathcal{L}_{\mathrm{reg}}
+\lambda_s\mathcal{L}_{\mathrm{smooth}}
\end{equation}
The four terms respectively enforce metric scene consistency, preserve the generated interaction structure, limit the residual magnitude, and maintain temporal smoothness.

Let $x_{t,i}^{\star}=x_i(S_t^{\star})$ denote the position of human body point $i$ induced by the refined state at frame $t$, and let $d_M(x)$ denote the signed distance from point $x$ to the registered scene $M$. In addition to $\mathcal{C}_h$, let $\mathcal{P}_h$ denote the set of human body points monitored for scene penetration. The metric-consistency term is:
\begin{equation}
\begin{aligned}
\mathcal{L}_{\mathrm{metric}}={}&
\sum_{(t,i)\in\mathcal{C}_h}
\rho\!\left(d_M(x_{t,i}^{\star})^2\right) \\
&+\beta_{\mathrm{pen}}
\sum_{t=1}^{T}\sum_{i\in\mathcal{P}_h}
\left[\max\!\left(0,-d_M(x_{t,i}^{\star})\right)\right]^2
\end{aligned}
\end{equation}
where $\rho(\cdot)$ is a robust penalty and $\beta_{\mathrm{pen}}\ge0$ controls the penetration penalty. The first term attracts intended interaction points toward the reconstructed scene surface, while the second penalizes body penetration.

Metric consistency alone can distort the original interaction structure. To preserve the generated human--scene relation, let $x_{t,i}^{\mathrm{HOI}}=x_i(S_t^{\mathrm{HOI}})$ denote the pre-refinement human point and let $y_{t,j}\in M$ be the associated scene point for edge $(i,j)\in\mathcal{E}_t$. Following interaction-mesh representations~\cite{omniretarget}, we define the pre-refinement relative offset as:
\begin{equation}
e^0_{t,ij}=x_{t,i}^{\mathrm{HOI}}-y_{t,j}
\end{equation}
and preserve it with $\mathcal{L}_{\mathrm{inter}}$:
\begin{equation}
\mathcal{L}_{\mathrm{inter}}
=
\sum_{t=1}^{T}\sum_{(i,j)\in\mathcal{E}_t}
\left\|
(x_{t,i}^{\star}-y_{t,j})-e^0_{t,ij}
\right\|_2^2
\end{equation}
The remaining two terms are:
\begin{align}
\mathcal{L}_{\mathrm{reg}}
&=\sum_{t=1}^{T}\|\delta S_t\|_{W_h}^2\\
\mathcal{L}_{\mathrm{smooth}}
&=\sum_{t=2}^{T}\|\delta S_t-\delta S_{t-1}\|_2^2
\end{align}

where $\|a\|_{W_h}^2=a^{\top}W_h a$ and $W_h$ is a diagonal positive-semidefinite human-state weighting matrix. The nonnegative weights $\lambda_m$, $\lambda_i$, $\lambda_r$, and $\lambda_s$ control the relative importance of the four terms. The optimal residual $\delta S_{1:T}^{\star}$ yields the refined interaction $S_{1:T}^{\star}$ through Eq.~(1), which is then passed to GMR.

\begin{figure*}[t]
\centering
\includegraphics[width=0.97\textwidth]{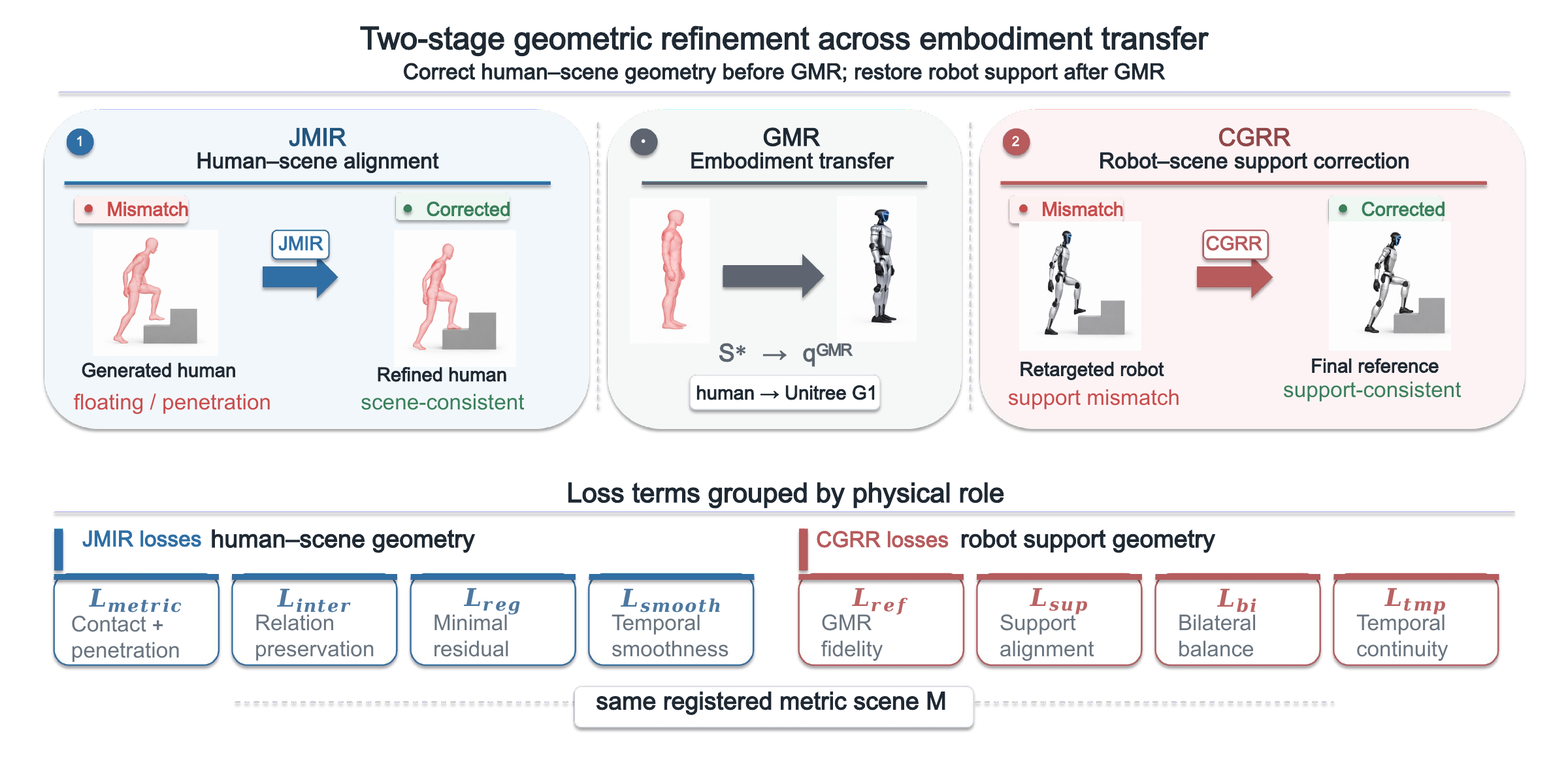}
\caption{Two-stage geometric refinement across embodiment transfer. JMIR corrects the human--scene geometric mismatch while preserving the generated interaction structure. After GMR, CGRR performs a bounded, contact-guided correction to recover robot-side support consistency.}
\label{fig:staged}
\end{figure*}

\subsubsection{Embodiment Transfer and Metric Support Representation}
\label{sec:support}
The refined human interaction is transferred to Unitree G1 using GMR~\cite{gmrpaper} as $q_t^{\mathrm{GMR}}=F_{\mathrm{GMR}}(S_t^{\star})$, where $F_{\mathrm{GMR}}$ denotes the human-to-G1 retargeting operator. Although $S_t^{\star}$ is geometrically consistent with the reconstructed scene, embodiment differences such as link lengths, sole geometry, joint limits, and kinematic structure can still introduce support mismatch after retargeting.

To quantify this robot--scene inconsistency, we construct a support height field $z_{\mathrm{sup}}(x,y)$ from the registered scene $M$. Let $\mathcal{F}=\{\ell,r\}$ denote the left and right feet. For foot $f\in\mathcal{F}$, let $z_f(q_t)$ denote the sole height and $p_f^{xy}(q_t)$ the horizontal position of its sole proxy. The signed support residual is:
\begin{equation}
r_f^z(q_t)
=
z_f(q_t)-z_{\mathrm{sup}}\!\left(p_f^{xy}(q_t)\right)-c
\end{equation}
where $c$ is the desired sole-clearance offset. A zero residual indicates alignment with the local support surface up to the prescribed clearance, while a nonzero residual measures the remaining vertical support mismatch.

\subsubsection{Robot--Scene Consistency Refinement}
\label{sec:cgrr}
CGRR corrects the robot--scene support mismatch through a bounded residual $\Delta q_t$ around the GMR reference. The optimization is solved sequentially over frames. At frame $t$, the previous optimized correction $\Delta q_{t-1}^{\star}$ is fixed and only the current correction $\Delta q_t$ is optimized. We formulate the per-frame problem as:
\begin{equation}
\Delta q_t^{\star}
=
\arg\min_{\Delta q_t}
\mathcal{L}_{\mathrm{CGRR}}^{t}
\end{equation}
subject to the feasibility constraints defined below, with
$
\mathcal{L}_{\mathrm{CGRR}}^{t}
=
\lambda_{\mathrm{ref}}\mathcal{L}_{\mathrm{ref}}^{t}
+\lambda_{\mathrm{sup}}\mathcal{L}_{\mathrm{sup}}^{t}
+\lambda_{\mathrm{bi}}\mathcal{L}_{\mathrm{bi}}^{t}
+\lambda_{\mathrm{tmp}}\mathcal{L}_{\mathrm{tmp}}^{t}
$.
Thus, the complete optimization is defined first, and the individual terms below specify how reference preservation, support correction, bilateral consistency, and temporal regularity are enforced.

The reference, support, and bilateral terms are:

\begin{align}
\mathcal{L}_{\mathrm{ref}}^{t}
&=\|\Delta q_t\|_{W_q}^2\\
\mathcal{L}_{\mathrm{sup}}^{t}
&=\sum_{f\in\mathcal{F}}
 w_{t,f}\left[r_f^z(q_t^{\mathrm{GMR}}+\Delta q_t)\right]^2\\
\mathcal{L}_{\mathrm{bi}}^{t}
&=\chi_t^{\mathrm{DS}}
\left[
 r_\ell^z(q_t^{\mathrm{GMR}}+\Delta q_t)
-r_r^z(q_t^{\mathrm{GMR}}+\Delta q_t)
\right]^2
\end{align}

and the temporal term is:
\begin{equation}
\mathcal{L}_{\mathrm{tmp}}^{t}
=
\begin{cases}
0, & t=1\\
\|\Delta q_t-\Delta q_{t-1}^{\star}\|_2^2, & t>1
\end{cases}
\end{equation}
Here $\|a\|_{W_q}^2=a^{\top}W_q a$, where $W_q$ is a diagonal positive-semidefinite residual-weight matrix. The nonnegative weights $\lambda_{\mathrm{ref}}$, $\lambda_{\mathrm{sup}}$, $\lambda_{\mathrm{bi}}$, and $\lambda_{\mathrm{tmp}}$ control the four terms. The reference term limits deviation from GMR, the support term reduces vertical error at active support feet, the bilateral term balances the two support residuals during double support, and the temporal term regularizes the current correction toward the previously optimized frame.

Support correction is applied only to feet that are likely to be in contact in the original GMR reference. We define $
w_{t,f}=\mathbb{I}\!\left(
|r_f^z(q_t^{\mathrm{GMR}})|<\tau_z
\;\wedge\;
\|v_{t,f}^{\mathrm{GMR}}\|_2<\tau_v
\right)
$,
where $v_{t,f}^{\mathrm{GMR}}$ is the Cartesian foot velocity evaluated from the GMR reference, $\tau_z$ is the support-distance threshold, $\tau_v$ is the foot-velocity threshold, and $\mathbb{I}(\cdot)$ is the indicator function. The double-support indicator is $\chi_t^{\mathrm{DS}}=w_{t,\ell}w_{t,r}$, so the bilateral term is active only when both feet satisfy the support criterion. This contact selection prevents CGRR from forcing swing feet toward the scene.

To keep the correction physically plausible and local to the GMR solution, the per-frame optimization is subject to
\begin{align}
\label{eq:cgrr_constraints}
& q_{\min} \le q_t^{\mathrm{GMR}}+\Delta q_t \le q_{\max}\\
& \|\Delta q_t\|_\infty \le \epsilon_q\\
& \|\Delta q_t\|_2 \le \Delta_{\mathrm{TR}}\\
& d_M\!\left(p_k(q_t^{\mathrm{GMR}}+\Delta q_t)\right)
  \ge -\epsilon_{\mathrm{pen}},\quad \forall k\in\mathcal{K}
\end{align}

Here $q_{\min}$ and $q_{\max}$ are the joint limits of the optimized variables, $\epsilon_q$ bounds the element-wise residual correction, and $\Delta_{\mathrm{TR}}$ is the trust-region radius. The set $\mathcal{K}$ contains robot points monitored for scene collision, $p_k(q)$ denotes the world position of query point $k$, and $\epsilon_{\mathrm{pen}}$ is the allowed penetration tolerance.
For a small correction, the support residual is locally linearized as:
\begin{equation}
r_f^z(q+\Delta q)
\approx
r_f^z(q)+J_f^z\Delta q
\end{equation}
where $J_f^z=\partial r_f^z(q)/\partial q$ is the support-residual Jacobian. We compute $J_f^z$ from the full kinematic chain using MuJoCo forward kinematics and restrict it to the selected correction variables. This local model is used by the CGRR solver to relate joint-space updates to support-residual changes.

Frames are processed sequentially from $t=1$ to $T$. Each constrained subproblem is solved using trust-region Sequential Quadratic Programming (SQP), with Operator Splitting Quadratic Program (OSQP) used for the quadratic subproblems. We use at most $K=12$ SQP updates per frame. The final robot reference is:
\begin{equation}
q_t^{\star}=q_t^{\mathrm{GMR}}+\Delta q_t^{\star}
\end{equation}
CGRR therefore preserves the overall GMR motion while applying only the bounded corrections required to restore robot--scene support consistency. Fig.~\ref{fig:staged} summarizes the complete two-stage refinement.

\subsection{Motion-Tracking Interface}
CGRR outputs a scene-consistent kinematic reference $q^{\star}_{1:T}$ rather than a controller-specific policy. The corresponding reference velocity $\dot q_t^{\star}$ is obtained from the optimized trajectory by finite differences. For tracking backend $b$, the reference is encoded as $r_t=\Phi_b\!\left(q_t^{\star},\dot q_t^{\star}\right)$, where $\Phi_b$ denotes the tracker-specific reference encoding. The control command is then produced by $a_t=\pi_{\theta_b}(o_t,r_t)$, where $o_t$ is the observation and $\pi_{\theta_b}$ denotes the policy of tracking backend $b$ with parameters $\theta_b$. Different tracking backends may use different reference encodings, policy architectures, and execution stacks while consuming the same scene-consistent reference generated by our pipeline. Our method therefore operates at the reference-generation level and does not require a controller-specific formulation of the refinement procedure.

\section{EXPERIMENTS}
We evaluate three research questions:       \textbf{RQ1:} Can the single-RGB-image workflow provide metric scene geometry and scene-consistent human interaction for humanoid reference generation? \textbf{RQ2:} Does two-stage refinement across embodiment transfer improve human--scene consistency before retargeting and robot--scene support consistency after retargeting? \textbf{RQ3:} Can the resulting references support physical Unitree G1 deployment across support-dominant terrains and multiple downstream tracking backends?

\noindent\textbf{Evaluation Protocol.}
All compared variants use the same reconstructed geometry, robot model, and evaluation criteria. The matched GRAIL comparison fixes the source interaction and reconstructed mesh. For the stage-aligned GRAIL--ours comparison, refinement time is measured on an NVIDIA RTX 4090; our reported 87~s includes JMIR and CGRR, with JMIR using $n_{\mathrm{iter}}=600$. Robot-side ablations use the same tracker and termination criteria over 100 simulation trials per variant. Tracker compatibility is evaluated by varying the tracking backend while fixing the generated robot reference.

\begin{table}[ht]
\caption{HOI comparison under a common evaluator. Refinement time is reported only for methods with a directly comparable scene-refinement stage; N/A denotes not applicable.}
\label{tab:hoi}
\centering
\small
\setlength{\tabcolsep}{2.8pt}
\resizebox{\columnwidth}{!}{%
\begin{tabular}{@{}lcccc@{}}
\toprule
Metric & HOI-Diff~\cite{hoidiff} & CHOIS~\cite{chois} & GRAIL~\cite{grail} & \textbf{Ours} \\
\midrule
Pen. Rate (\%)$\downarrow$      & 3.04   & 4.15   & 1.59   & \textbf{1.23} \\
Interaction$\uparrow$          & 1.56   & 2.46   & 3.57   & \textbf{3.95} \\
Extreme Frame (\%)$\downarrow$ & 5.38   & 6.32   & 3.46   & \textbf{1.45} \\
H.Smo.$\downarrow$             & 0.0065 & 0.0053 & 0.0035 & \textbf{0.0031} \\
Body Dev.$\downarrow$          & 0.251  & 0.264  & 0.126  & \textbf{0.106} \\
InterMimic SR (\%)$\uparrow$    & 19.4   & 12.3   & 63.5   & \textbf{84.3} \\
Refinement Time (s)$\downarrow$ & N/A    & N/A    & 480    & \textbf{87} \\
\bottomrule
\end{tabular}%
}
\end{table}
\medskip
\noindent\textbf{RQ1: Scene and Interaction Quality.}
As an evaluation-only metric-scale sanity check on the Box20 scene, the physical height, length, and width are 20, 100, and 50~cm, while the reconstructed mesh gives 19.4, 97.6, and 48.8~cm, respectively, corresponding to a 2.6\% mean relative dimensional error. The physical dimensions are used only for evaluation and are never provided to reconstruction or refinement.

Table~\ref{tab:hoi} evaluates interaction quality before robot-side execution. Relative to GRAIL, ours reduces penetration from 1.59\% to 1.23\% and increases the interaction score from 3.57 to 3.95. All baselines use the same evaluator and metric definitions, but not necessarily identical inputs. HOI-Diff and CHOIS are generation-based methods whose pipelines do not include a directly comparable scene-refinement stage under our evaluation protocol; therefore, refinement time is reported as N/A for these methods. For the stage-aligned GRAIL--ours comparison, refinement time decreases from 480~s to 87~s, yielding a 5.5$\times$ speedup.
\begin{table}[hp]
\caption{Robot-side retargeting comparison and staged-refinement ablation over 100 complete simulation trials per method.}
\label{tab:robot}
\centering
\footnotesize
\begin{tabular*}{\columnwidth}{@{\extracolsep{\fill}}lccc@{}}
\toprule
Method / Variant & \shortstack{SupportErr\\(cm)$\downarrow$} & \shortstack{MPJPE-L\\(mm)$\downarrow$} & \shortstack{SR\\(\%)$\uparrow$} \\
\midrule
PHC-based reference~\cite{phc} & 0.74 & 59.8 & 69 \\
GMR~\cite{gmrpaper,gmrrepo} & 0.57 & 50.7 & 75 \\
JMIR + GMR & 0.46 & 46.8 & 80 \\
GMR + root-only & 0.76 & 43.6 & 76 \\
GMR + CGRR w/o bilateral & 0.49 & 42.6 & 82 \\
GMR + full CGRR & 0.24 & 40.3 & 88 \\
\textbf{JMIR + GMR + CGRR} & \textbf{0.21} & \textbf{38.2} & \textbf{89} \\
\bottomrule
\end{tabular*}
\end{table}
\medskip

\begin{figure}[!hp]
\centering
\includegraphics[height=4.2cm, width=0.99\columnwidth]{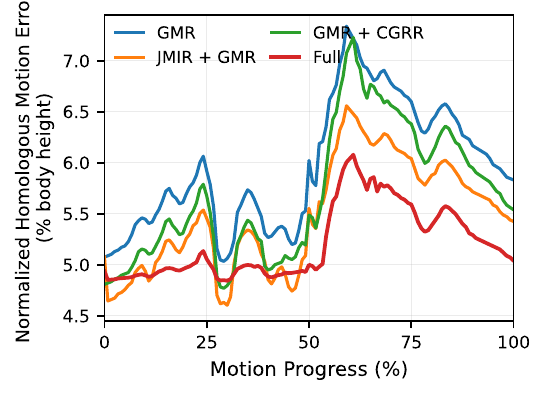}
\caption{Aggregated source-motion fidelity over normalized motion progress. Sequences are temporally normalized from 0\% to 100\% before aggregation; lower normalized homologous motion error indicates better fidelity.}
\label{fig:motion-fidelity}
\end{figure}

\noindent\textbf{RQ2: Two-Stage Consistency Refinement.}
We additionally assess source-motion fidelity using the normalized homologous motion error shown in Fig.~\ref{fig:motion-fidelity}, where lower values indicate better preservation of the source motion. To aggregate motions of different durations, each sequence is temporally normalized to a motion progress from 0\% to 100\% before the fidelity curves are computed over the evaluated motion set. Following the GRAIL evaluation convention, each evaluated variant $m$ produces a kinematic robot reference $\{\mathbf{q}_{t}^{\mathrm{ref},m}\}_{t=1}^{T}$ that is tracked by the same downstream controller in simulation. We evaluate reference geometry and tracking behavior using SupportErr, local Mean Per-Joint Position Error (MPJPE-L), and simulation Success Rate (SR).

SupportErr measures the mean absolute sole--support residual of the kinematic reference over active support frames
\begin{equation}
E_{\mathrm{sup}}^{m}
=
\frac{
\displaystyle\sum_{t=1}^{T}\sum_{f\in\mathcal{F}}
 w_{t,f}^{m}
 \left|r_f^z\!\left(\mathbf{q}_{t}^{\mathrm{ref},m}\right)\right|
}{
\displaystyle\sum_{t=1}^{T}\sum_{f\in\mathcal{F}}w_{t,f}^{m}
}
\end{equation}
Here $m$ indexes the evaluated retargeting variant, $\mathbf{q}_{t}^{\mathrm{ref},m}$ is the robot configuration generated by variant $m$ at frame $t$, and $w_{t,f}^{m}$ is the active-support indicator obtained by applying the criterion to that reference. The frame count $T$ and foot set $\mathcal{F}$ are defined in Sec.~III, while $r_f^z(\cdot)$  is the signed support residual. SupportErr is reported in centimeters.
\begin{figure}[!htb]
\centering
\includegraphics[
    width=0.8\columnwidth,
    height=0.55\textheight,
    keepaspectratio
]{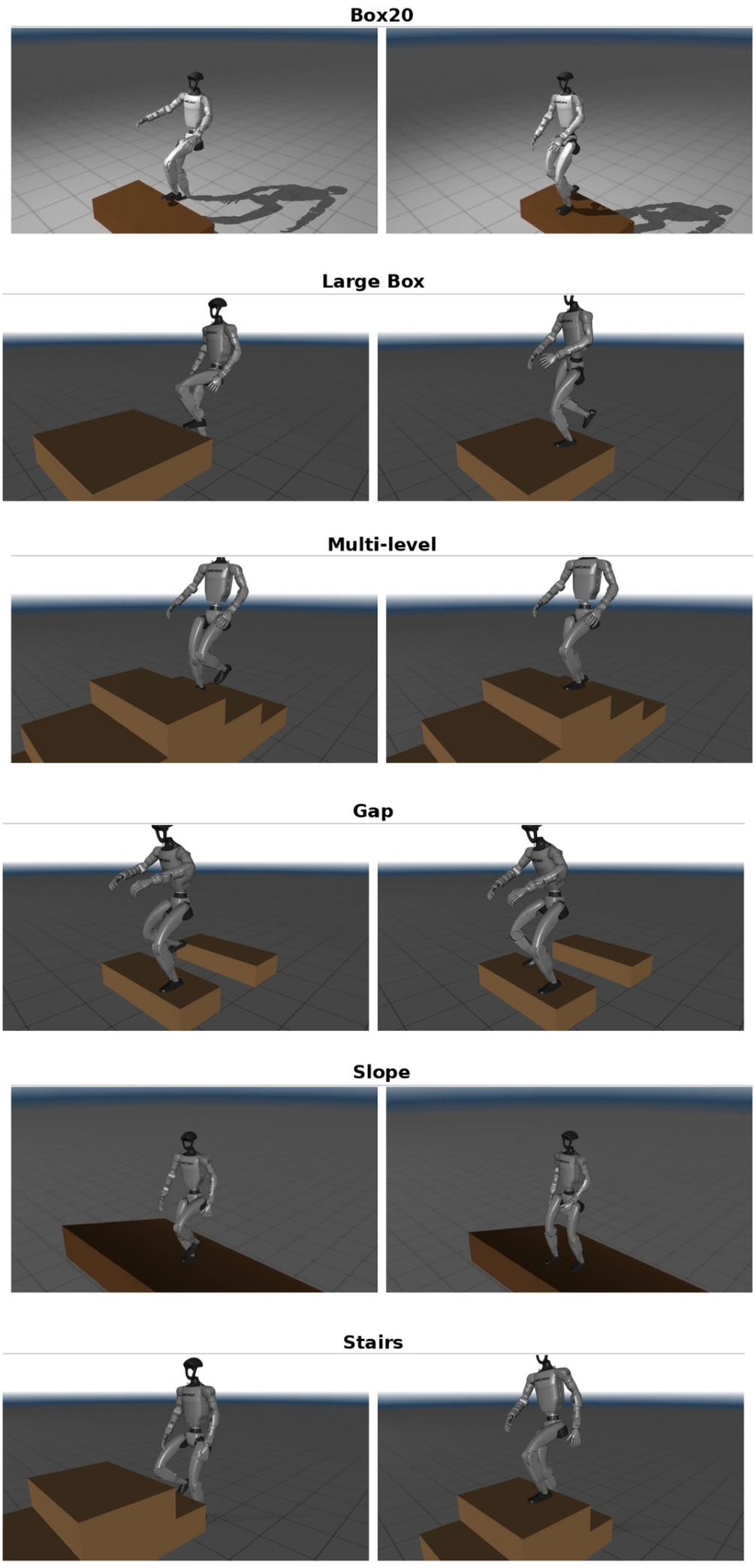}
\caption{Representative simulation executions across six support-dominant scenes. Rows correspond to Box 20 cm, Large Box, Multi-level, Gap, Slope, and Stairs; columns show two frames per scene.}
\label{fig:sim}
\end{figure}

MPJPE-L measures the discrepancy between the simulated robot motion and its corresponding kinematic reference in root-relative coordinates. Let $\mathcal{J}$ be the body-link set used by the evaluator. For link $j\in\mathcal{J}$, define
\begin{align}
\widetilde{\mathbf{p}}_{t,j}^{\mathrm{sim},m}
&=
\left(\mathbf{R}_{t,\mathrm{root}}^{\mathrm{sim},m}\right)^{\top}
\left(\mathbf{p}_{t,j}^{\mathrm{sim},m}-\mathbf{p}_{t,\mathrm{root}}^{\mathrm{sim},m}\right) \\
\widetilde{\mathbf{p}}_{t,j}^{\mathrm{ref},m}
&=
\left(\mathbf{R}_{t,\mathrm{root}}^{\mathrm{ref},m}\right)^{\top}
\left(\mathbf{p}_{t,j}^{\mathrm{ref},m}-\mathbf{p}_{t,\mathrm{root}}^{\mathrm{ref},m}\right)
\end{align}

where $\mathbf{p}_{t,j}^{\mathrm{sim},m},\mathbf{p}_{t,j}^{\mathrm{ref},m}\in\mathbb{R}^{3}$ are the world-frame positions of link $j$ in the simulated robot and its kinematic reference. The vectors $\mathbf{p}_{t,\mathrm{root}}^{\mathrm{sim},m}$ and $\mathbf{p}_{t,\mathrm{root}}^{\mathrm{ref},m}$ are the corresponding root positions, while $\mathbf{R}_{t,\mathrm{root}}^{\mathrm{sim},m},\mathbf{R}_{t,\mathrm{root}}^{\mathrm{ref},m}\in SO(3)$ are the corresponding root rotation matrices. MPJPE-L is
\begin{equation}
E_{\mathrm{MPJPE-L}}^{m}
=
\frac{1}{|\mathcal{T}_{\mathrm{eval}}^{m}|\,|\mathcal{J}|}
\sum_{t\in\mathcal{T}_{\mathrm{eval}}^{m}}
\sum_{j\in\mathcal{J}}
\left\|
\widetilde{\mathbf{p}}_{t,j}^{\mathrm{sim},m}
-
\widetilde{\mathbf{p}}_{t,j}^{\mathrm{ref},m}
\right\|_2
\end{equation}
Here $\mathcal{T}_{\mathrm{eval}}^{m}$ is the set of simulation frames retained by the evaluator, $|\cdot|$ denotes set cardinality, and $\|\cdot\|_2$ is the Euclidean norm. MPJPE-L is reported in millimeters. Simulation SR is $100N_{\mathrm{succ}}^{m}/N_{\mathrm{trial}}^{m}$ percent, where $N_{\mathrm{succ}}^{m}$ is the number of successfully completed trials and $N_{\mathrm{trial}}^{m}=100$ is the total number of trials for variant $m$.
\begin{figure*}[!t]
\centering
\includegraphics[
    width=0.90\textwidth,
    height=0.38\textheight,
    keepaspectratio
]{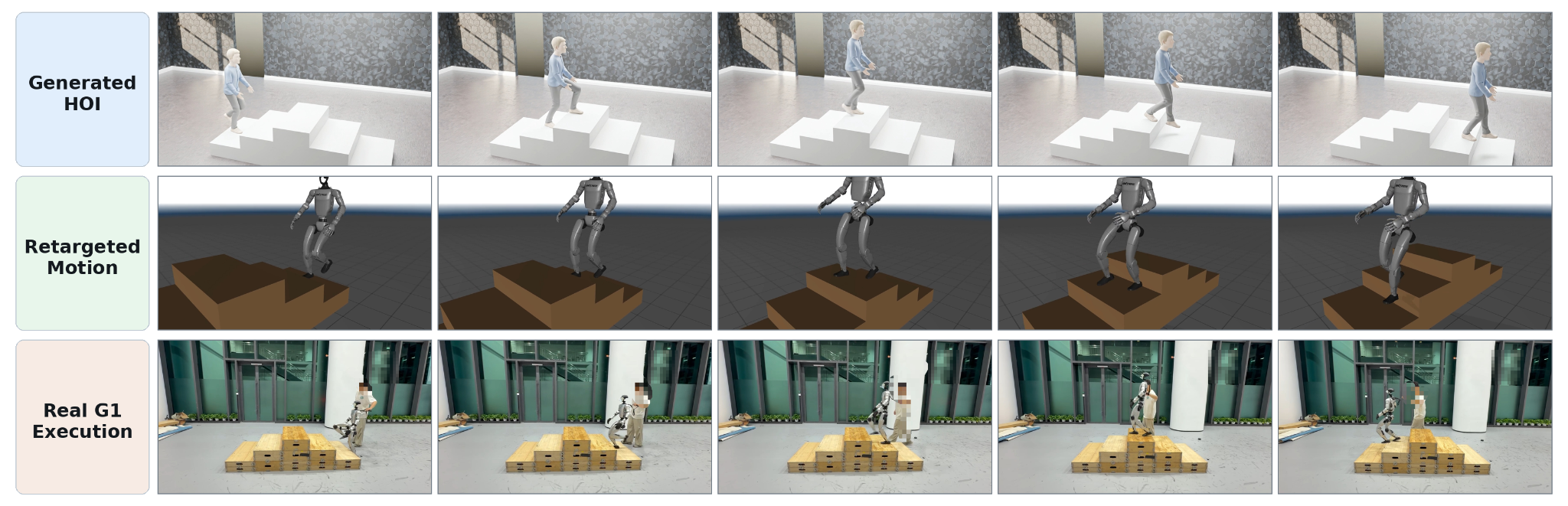}
\caption{End-to-end qualitative sequence from human interaction generation to physical Unitree G1 execution. Rows show the generated HOI, the refined robot reference after retargeting, and physical G1 execution; columns show successive frames of the same support-dominant interaction.}
\label{fig:real}
\end{figure*}

As shown in Table~\ref{tab:robot}, adding JMIR before GMR improves SupportErr from 0.57 to 0.46~cm, MPJPE-L from 50.7 to 46.8~mm, and simulation SR from 75\% to 80\%. CGRR further corrects the post-retargeting support mismatch. The complete JMIR--GMR--CGRR pipeline obtains 0.21~cm SupportErr, 38.2~mm MPJPE-L, and 89\% simulation SR. Representative simulation executions across the six support-dominant settings are shown in Fig.~\ref{fig:sim}.

\begin{figure}[htbp]
\centering
\includegraphics[height=4.8cm, width=0.96\columnwidth]{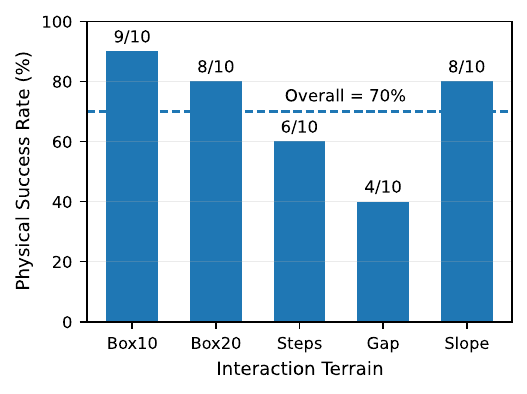}
\caption{Terrain-wise physical execution success on Unitree G1. Each terrain is evaluated over 10 real-robot trials; the dashed line denotes the overall success rate of 70\% across 50 trials.}
\label{fig:terrain-success}
\end{figure}
\medskip
\noindent\textbf{RQ3: Real-Robot Deployment and Tracker Compatibility.}
The end-to-end progression from generated HOI to retargeted reference
and physical execution is illustrated in Fig.~\ref{fig:real}.
Each terrain is evaluated over 10 complete Unitree G1 trials.
A trial is successful when the robot completes the prescribed
interaction without a fall or emergency termination.
The physical interaction geometry follows the reconstructed mesh
derived from the RGB input.

As shown in Fig.~\ref{fig:terrain-success}, across the five terrain categories, the complete workflow succeeds in 35/50 trials, corresponding to a 70\% physical success rate. The result demonstrates physical execution across multiple support-dominant reconstructed scenes.

\noindent\textit{Tracker compatibility.}
Table~\ref{tab:tracker} compares downstream tracker compatibility and scene-specific adaptation cost. We keep the generated robot reference fixed and change only the downstream tracking backend. Direct settings use pretrained policies without scene-specific fine-tuning. We additionally report a task-specific fine-tuned SONIC configuration (SONIC-FT) to characterize the performance/adaptation-cost trade-off.
\begin{table}[htp]
\caption{Tracker compatibility and scene-specific adaptation cost.}
\label{tab:tracker}
\centering
\footnotesize
\begin{tabular*}{\columnwidth}{@{\extracolsep{\fill}}lcccc@{}}
\toprule
Tracker & Fine-tune & Trials & SR (\%)$\uparrow$ & Prep./FT time \\
\midrule
Humanoid-GPT~\cite{humanoidgpt} & No & 10 & 50 & $<$1 min \\
MimicLite~\cite{mimiclite} & No & 10 & 70 & $<$1 min \\
GMT~\cite{gmt} & No & 10 & 60 & $<$1 min \\
SONIC~\cite{sonic} & No & 10 & 60 & $<$1 min \\
\midrule
SONIC-FT~\cite{sonic} & Yes & 10 & \textbf{90} & 30 h \\
\bottomrule
\end{tabular*}
\end{table}

The fixed generated reference can be consumed by several independently developed trackers. Task-specific SONIC adaptation increases SR to 90\% at a 30~h adaptation cost. The $<$1~min entries denote tracker-specific preparation after the robot reference has already been generated and exclude the 87~s reference-refinement runtime.

\FloatBarrier

\section{CONCLUSION}
We presented a single-RGB-image-to-executable-humanoid workflow for support-dominant scene interaction that connects scene reconstruction, generative HOI synthesis, GMR retargeting, and robot-side geometric refinement. The central design separates two sources of inconsistency across embodiment transfer: JMIR corrects human--scene mismatch before GMR, whereas CGRR corrects robot-specific support mismatch introduced after transfer. The proposed workflow improves interaction quality and reduces refinement time relative to GRAIL. The complete JMIR--GMR--CGRR pipeline obtains a support error of 0.21~cm, an MPJPE-L of 38.2~mm, and a simulation success rate of 89\%; across 50 Unitree G1 trials it achieves a 70\% physical success rate. The resulting motion references are consistent with the scene geometry, enabling physical execution across support-dominant interactions and integration with multiple humanoid motion trackers.

Future work will extend the pipeline beyond support-dominant interactions to tasks involving dynamic object manipulation and changing contact conditions. Incorporating dynamic feasibility into reference refinement could further improve physical execution, while accounting for uncertainty in scene reconstruction could make contact alignment more robust to geometric errors. How execution feedback can guide the adaptation of motion references to differences between the reconstructed scene and the physical environment is also an interesting direction for future research.


\begin{thebibliography}{22}
\bibitem{guo} C. Guo, S. Zou, X. Zuo, S. Wang, W. Ji, X. Li, and L. Cheng, ``Generating diverse and natural 3D human motions from text,'' in \emph{Proc. IEEE/CVF Conf. Comput. Vis. Pattern Recognit. (CVPR)}, 2022, pp. 5152--5161.
\bibitem{mdm} G. Tevet, S. Raab, B. Gordon, Y. Shafir, D. Cohen-Or, and A. H. Bermano, ``Human motion diffusion model,'' in \emph{Proc. Int. Conf. Learn. Represent. (ICLR)}, 2023.
\bibitem{grail} T. Xie, H. Zhang, J. Park, Z. Wang, B. Wen, J. Li, X. Li, Q. Ben, H. Weng, Y. Ye, D. Minor, T. Wang, C. Jiang, S. Fidler, J. Kautz, L. Fan, Y. Zhu, Z. Luo, U. Iqbal, and Y. Yuan, ``GRAIL: Generating humanoid loco-manipulation from 3D assets and video priors,'' arXiv:2606.05160, 2026.
\bibitem{hoidiff} X. Peng, Y. Xie, Z. Wu, V. Jampani, D. Sun, and H. Jiang, ``HOI-Diff: Text-driven synthesis of 3D human-object interactions using diffusion models,'' in \emph{CVPR Workshop on Human Motion Generation}, 2025.
\bibitem{chois} J. Li, A. Clegg, R. Mottaghi, J. Wu, X. Puig, and C. K. Liu, ``Controllable human-object interaction synthesis,'' in \emph{Proc. Eur. Conf. Comput. Vis. (ECCV)}, 2024.
\bibitem{david} H. Kim, S. Baik, and H. Joo, ``DAViD: Modeling dynamic affordance of 3D objects using pre-trained video diffusion models,'' in \emph{Proc. IEEE/CVF Int. Conf. Comput. Vis. (ICCV)}, 2025, pp. 10330--10341.
\bibitem{sam3d} X. Chen, F.-J. Chu, P. Gleize, K. J. Liang, A. Sax, H. Tang, W. Wang, M. Guo, T. Hardin, X. Li, et al., ``SAM 3D: 3Dfy anything in images,'' in \emph{Proc. IEEE/CVF Conf. Comput. Vis. Pattern Recognit. (CVPR)}, 2026, pp. 7220--7232.
\bibitem{phc} Z. Luo, J. Cao, A. W. Winkler, K. Kitani, and W. Xu, ``Perpetual humanoid control for real-time simulated avatars,'' in \emph{Proc. IEEE/CVF Int. Conf. Comput. Vis. (ICCV)}, 2023.
\bibitem{gmrpaper} J. P. Araujo, Y. Ze, P. Xu, J. Wu, and C. K. Liu, ``Retargeting matters: General motion retargeting for humanoid motion tracking,'' arXiv:2510.02252, 2025.
\bibitem{gmrrepo} Y. Ze, J. P. Araujo, J. Wu, and C. K. Liu, ``GMR: General motion retargeting,'' 2025. [Online]. Available: \url{https://github.com/YanjieZe/GMR}
\bibitem{omniretarget} L. Yang, X. Huang, Z. Wu, A. Kanazawa, P. Abbeel, C. Sferrazza, C. K. Liu, R. Duan, and G. Shi, ``OmniRetarget: Interaction-preserving data generation for humanoid whole-body loco-manipulation and scene interaction,'' arXiv:2509.26633, 2025.
\bibitem{kinodynamic} X. Zhang, S. Haener, V. Madabushi, and M. Tucker, ``Kinodynamic motion retargeting for humanoid locomotion via multi-contact whole-body trajectory optimization,'' arXiv:2603.09956, 2026.
\bibitem{resmimic} S. Zhao, Y. Ze, Y. Wang, C. K. Liu, P. Abbeel, G. Shi, and R. Duan, ``ResMimic: From general motion tracking to humanoid whole-body loco-manipulation via residual learning,'' arXiv:2510.05070, 2025.
\bibitem{videomimic} A. Allshire, H. Choi, J. Zhang, D. McAllister, A. Zhang, C. M. Kim, T. Darrell, P. Abbeel, J. Malik, and A. Kanazawa, ``Visual imitation enables contextual humanoid control,'' in \emph{Proc. Conf. Robot Learn. (CoRL)}, vol. 305, 2025, pp. 794--815.
\newpage
\bibitem{beyondmimic} Q. Liao, T. E. Truong, X. Huang, G. Tevet, K. Sreenath, and C. K. Liu, ``BeyondMimic: From motion tracking to versatile humanoid control via guided diffusion,'' arXiv:2508.08241, 2025.
\bibitem{gmt} Z. Chen, M. Ji, X. Cheng, X. Peng, X. B. Peng, and X. Wang, ``GMT: General motion tracking for humanoid whole-body control,'' arXiv:2506.14770, 2025.
\bibitem{sonic} Z. Luo, Y. Yuan, T. Wang, C. Li, S. Chen, F. Castaneda, Z.-A. Cao, J. Li, D. Minor, Q. Ben, et al., ``SONIC: Supersizing motion tracking for natural humanoid whole-body control,'' arXiv:2511.07820, 2025.
\bibitem{intermimic} S. Xu, H. Y. Ling, Y.-X. Wang, and L.-Y. Gui, ``InterMimic: Towards universal whole-body control for physics-based human-object interactions,'' in \emph{Proc. IEEE/CVF Conf. Comput. Vis. Pattern Recognit. (CVPR)}, 2025, pp. 12266--12277.
\bibitem{scenebot} S. Chen, S. Zhao, Z. Wu, J. Li, G. Shi, and C. K. Liu, ``SceneBot: Contact-prompted general humanoid whole body tracking with scene-interaction,'' arXiv:2606.27581, 2026.
\bibitem{hiking} S. Zhu, Z. Zhuang, M. Zhao, K.-Y. Lee, and H. Zhao, ``Hiking in the Wild: A scalable perceptive parkour framework for humanoids,'' arXiv:2601.07718, 2026.
\bibitem{mimiclite} RoboParty Lab Team, ``MimicLite: Efficient and effective general humanoid motion tracking,'' Technical Report, 2026. [Online]. Available: \url{https://github.com/Roboparty/MimicLite}
\bibitem{humanoidgpt} Z. Qi, X. Chen, D. Liu, C. Lin, Y. Lian, S. Liang, Z. Zhang, Y. Guan, J. Wang, W. Zhang, X. Yu, H. Wang, and L. Yi, ``Humanoid-GPT: Scaling data and structure for zero-shot motion tracking,'' in \emph{Proc. IEEE/CVF Conf. Comput. Vis. Pattern Recognit. (CVPR)}, 2026.
\end{thebibliography}
\end{document}